\documentclass[sigconf]{aamas} 

\usepackage{balance} % for balancing columns on the final page

\usepackage{pgfplots}
\usepackage{tikz}
\usetikzlibrary{arrows,shadows,positioning}
\usepackage{float}

\usepackage{graphicx}
\usepackage{subfig}

\usepackage{algorithm}
\usepackage{algpseudocode}

\tikzset{
  frame/.style={
    rectangle, draw,
    text width=6em, text centered,
    minimum height=4em,drop shadow,fill=white,
    rounded corners,
  },
  line/.style={
    draw, -latex',rounded corners=3mm,
  }
}

\setcopyright{none}
\acmConference[ALA '23]{Proc.\@ of the Adaptive and Learning Agents Workshop (ALA 2023)}
{May 29-30, 2023}{London, UK, \url{https://alaworkshop2023.github.io/}}{Cruz, Hayes, Wang, Yates (eds.)}
\copyrightyear{2023}
\acmYear{2023}
\acmDOI{}
\acmPrice{}
\acmISBN{}
\title[AAMAS-2023 Formatting Instructions]{Evolutionary Strategy Guided Reinforcement Learning via MultiBuffer Communication}

\author{Adam Callaghan}
\affiliation{
  \institution{University of Galway}
  \city{Galway}
  \country{Ireland}}
\email{A.Callaghan10@nuigalway.ie}

\author{Karl Mason}
\affiliation{
  \institution{University of Galway}
  \city{Galway}
  \country{Ireland}}
\email{karl.mason@universityofgalway.ie}

\author{Patrick Mannion}
\affiliation{
  \institution{University of Galway}
  \city{Galway}
  \country{Ireland}}
\email{patrick.mannion@universityofgalway.ie}

\begin{abstract}
Evolutionary Algorithms and Deep Reinforcement Learning have both successfully solved control problems across a variety of domains.  Recently, algorithms have been proposed which combine these two methods, aiming to leverage the strengths and mitigate the weaknesses of both approaches. 

A central component of algorithms that combine Evolutionary Algorithms with Deep Reinforcement Learning has been the "Shared Replay Buffer". Deep Reinforcement Learning algorithms require batches of data to update policy networks. Since Evolutionary Algorithms encounter such data in excess, they can feed the data produced from a variety of different behavioural policies to the Deep Reinforcement Learning model. Deep Reinforcement Learning in-turn seeks to bias the Evolutionary Algorithm to higher areas of fitness by introducing high-performing individuals into the population periodically. This paradigm has produced several highly successful algorithms. 

In this paper we introduce a new Evolutionary Reinforcement Learning model built on this framework, combining a particular family of Evolutionary algorithm called Evolutionary Strategies with the off-policy Deep Reinforcement Learning algorithm TD3. The framework utilises a multi-buffer system instead of using a single shared replay buffer. The multi-buffer system allows for the Evolutionary Strategy to search freely in the search space of policies, without running the risk of overpopulating the replay buffer with poorly performing trajectories which limit the number of desirable policy behaviour examples thus negatively impacting the potential of the Deep Reinforcement Learning within the shared framework.

The proposed algorithm is demonstrated to perform competitively with current Evolutionary Reinforcement Learning algorithms on MuJoCo control tasks, outperforming the well known state-of-the-art CEM-RL on 3 of the 4 environments tested.
\end{abstract}


\newcommand{\BibTeX}{\rm B\kern-.05em{\sc i\kern-.025em b}\kern-.08em\TeX}

\pgfplotsset{compat=1.18}
\begin{document}

\pagestyle{fancy}
\fancyhead{}

\maketitle 

\section{Introduction}
Reinforcement Learning (RL) achieved notable success over the last few decades, from playing board games like chess \cite{Tesauro1995} and GO! \cite{Silver2016} at near human-expert levels, playing video games where the agent is fed pixel inputs of the game \cite{Mnih2015}, to self driving cars \cite{Lv2022}.

Evolutionary Algorithms (EAs) have existed concurrently, with deep roots in function optimisation problems.  EAs seek to find the optimal set of parameters (often belonging to some function-approximator) in order to minimise the a loss function.  EAs have been applied to a variety of real world problems such as: video game level generation \cite{Jiang2021}, determining an optimal architecture for a function approximator \cite{Stanley2001} and also the control problem of finding a behavioural policy that allows an agent to perform a task sufficiently \cite{Segal2022}.

In general an EA maintains a population of one or more individuals.  In the case of finding an optimal control policy, an individual or agent is usually represented by the weight vector of its policy Neural Network.  This Neural Network maps states to actions, and is used to determine the actions the agent performs in the environment.  EA make use of genetic operators such as mutation and recombination to apply a selective pressure to the population leading to the emergence of fitter agents.  In practice, knowing the fitness of the entire population at any point in time is a must. This results in many evaluations of the policy networks causing "low sample efficiency".  

Conversely, Deep Reinforcement Learning (DRL) generally consists of a single agent again represented as one (or in the case of actor-critic algorithms - two) Neural Networks.  The performance of this agent is improved upon by performing Stochastic Gradient Descent on the Neural Networks using batches \cite{Mnih2015} of $(s,a,r,s')$ data points, where $a$ is the action that was taken when the environment was in state $s$, resulting in the environment transitioning to state $s'$ and receiving back a scalar reward $r$.  The updates performed on the Neural Networks, either directly or indirectly, seek to increase the total reward gathered by the policy described by the networks.  Since data points can be reused it leads to a higher sample efficiency than EAs.  However, in practice DRL algorithms have been observed to be more fragile to hyper-parameter choice as seen in \cite{Franceschetti2022} where the difference between setting a particular hyper-parameter to 0.9999 instead of 0.99 can be the difference between the agent solving the problem completely or not at all.

Algorithms have been proposed which seek to unite these two frameworks while preserving the desirable traits of both \cite{sigaud2022combining}.  Most notably and related to the work in this paper, is the algorithm titled ERL \cite{Khadka2018}, where a general framework is described. The framework can be summarised as follows: A Genetic Algorithm containing k agents and a single Reinforcement Learner are initialised.  The agents in the Genetic Algorithm are evaluated and by means of tournament selection \cite{mitchell1997machine} the fitter agents are assigned a higher probability of passing their genes through to the next generation. The evaluation trajectories of the GA are stored in a replay buffer which is shared with the trajectories created by the RL agent.  As such, the batches sampled from this replay buffer by the RL agent, come from a range of different behavioural policies.  The Reinforcement Learner is periodically allowed to overwrite the weakest member of the GA, in hopes of biasing the GA towards regions of search space with higher fitness.  In turn, this allows the data the GA generates and feeds to the Reinforcement Learner to be of higher quality.

This paper proposes a new algorithm inspired by ERL but achieving higher results through two key differences. 
\begin{itemize}
  \item Firstly, GAs have been observed to cause catastrophic forgetting when used in control problems \cite{Bodnar2020}. As such we investigate another family of EAs - Evolutionary Strategies.  In particular, we apply the algorithm titled ES (sometimes referred to as openES to avoid confusion with the family of algorithms - ES). \cite{Salimans2017}
  \item Secondly, we propose that the use of multiple replay buffers can increase the performance of this algorithm.  This is founded on the claim that EAs often generate new individuals by randomly sampling from nearby policies.  Since the fitness landscapes of many control problems are often extremely non-smooth \cite{Sullivan2022} it is unwise to assume that every policy near to a "good" policy will in itself be "good".  This can lead to a single replay buffer being unbalanced with respect to good and bad trajectories.  A simple solution would be to compartmentalise the buffer into "good"/"bad" trajectories and sampling batches according to some desired ratio.
  In this paper a third buffer is also used where the Reinforcement Learner can store its own exploratory sequences.  

\end{itemize}

%%%%%%%%%%%%%%%%%%%%%%%%%%%%%%%%%%%%%%%%%%%%%%%%%%%%%%%%%%%%%%%%%%%%%%%%

\section{Background and Related Works}
%%%% write about MDP - TD3, ES, CEM, CEM-RL, AES-RL

\textbf{Markov Decision Process:} A Markov Decision Process (MDP) is used to convert a sequential control task into a mathematical formulation.  MDPs are represented as the tuple (\textit{S},\textit{A},\textit{T} ,R,$\gamma$).  \textit{S} is the set of all states the environment can be in.  In the case of playing board games, \textit{S} would be the set of every possible configuration the pieces can be in.  \textit{A} is the set of all actions the agent may take in a state.  \textit{T} is the transition dynamics of the environment.  It is a map $\textit{S} : (\textit{S},\textit{A},\textit{S}) \rightarrow (0,1)$ giving the probability that taking an action in a state will result in the environment changing to another particular state.  R is the reward function, which gives the reward $r$ received by the agent at each timestep after performing an action.  It is usually assumed that these rewards take scalar values. $\gamma$ is a hyper-parameter which takes values in the range $[0,1]$.  Its role is to balance the agent's interests towards acting to receive maximal rewards over a long period of time (when $\gamma=1$) or acting to maximise immediate rewards (when $\gamma=0$).  

Dynamic Programming \cite{Sutton1998} is used to find the behavioural policies that maximise the agents cumulative rewards - often simply called the optimal policy.  

\textbf{Reinforcement Learning:}  When the transition dynamics \textit{T} are unknown, dynamic programming cannot be used, instead RL may be used to learn a policy.  RL algorithms can be model-based \cite{Sutton1998} in that they learn an approximation of the dynamics and find an optimal policy based on this, or model-free where they learn an optimal policy by acting through trial and error in the environment. In this paper we focus on model-free algorithms.  

Model-free RL can be divided further into on-policy and off-policy.  In on-policy RL, the agent must learn from batches of data generated by action selections under its own policy $\pi$.  On the other hand, off-policy RL is free to learn from batches of data collected from a completely different behavioural policy than the one it is currently following.  

Off-policy RL allows for an agent to store trajectories it generates in a Replay Buffer \cite{Mnih2015}, which it can then sample batches from.  The use of a replay buffer helps to increase sample efficiency massively while additionally helping protect the agent from forgetting prior learned traits - referred to as catastrophic forgetting.

The most widely used RL algorithms today are off-policy including DQN, TD3 \cite{Fujimoto2018} (an improvement on DDPG \cite{Lillicrap2015} reducing the agent overestimating the value of actions) and SAC \cite{Haarnoja2018}.  The common representations of agents in RL include value-based agents which approximate the expected cumulative discounted rewards associated to each state-pair and define the policy based on this,  actor-based agents who learn the policy without the need of value estimates and finally actor-critic methods which approximate the value and in turn use this information to aid the learning of the policy.

\textbf{Evolutionary Strategies:}  Evolutionary Strategies are one of three main sub-families of Evolutionary Algorithms, along with Genetic Algorithms and Genetic Programming.  While Evolutionary Strategies are often applied to numerical optimisation problems, here we discuss how they can be applied to solving control problems. An ES is instantiated by placing a parameterised probability distribution over the search space - in this case the space of all policy network parameters.  While a range of probability distributions could be selected, previous studies have demonstrated that using a multivariate isotropic Gaussian with mean parameter $\theta$ and standard deviation $\sigma$ allows for significant run-time speed-ups \cite{Salimans2017}.  Under this configuration, the algorithm holds fixed the standard deviation of the distribution but performs approximate gradient ascent on the mean parameter with respect to the policy's fitness function $J(\theta)$ by evaluating the fitness of policies selected through the distribution in a method that closely resembles finite-differencing. 

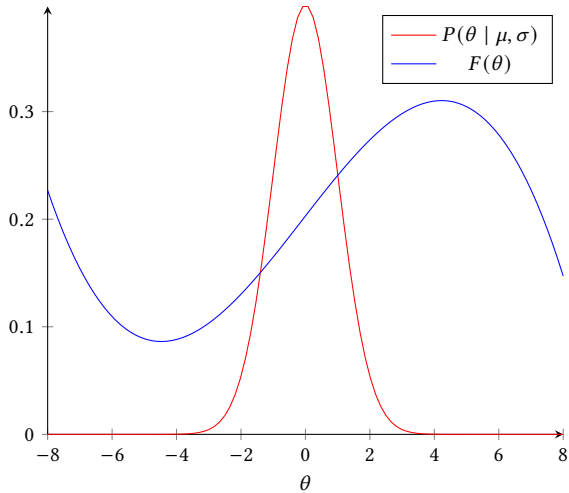
\begin{figure}[h]
  \centering
  \small
  \begin{tikzpicture}
  \begin{axis}[
      axis lines = left,
      xlabel = \(\theta\),
      ]
   %Below the red parabola is defined
  \addplot [
  domain=-8:8, 
  samples=100, 
  color=red,
  ]
  {(1/sqrt(2*pi))*e^((-1/2)*(x)^2)};
  \addlegendentry{$P(\theta \mid \mu, \sigma)$}

  \addplot [
  domain=-8:8, 
  samples=100, 
  color=blue,
  ]
  {0.203 + 0.0386*x -0.00025*x^2 - 0.000681*x^3};
  \addlegendentry{$F(\theta)$}

  \end{axis}
  \end{tikzpicture}
  \caption{Evolutionary Strategies}{The red curve represents a Gaussian probability density function (PDF) with mean $\mu$ and standard deviation $\sigma$ over a 1D parameter space.  The blue curve represents the fitness of the policy defined by the weights $\theta$.  In control problems this fitness function is unknown but can be sampled.  ES applies approximate gradient ascent to the mean, effectively "sliding" the distribution along the x-axis until it finds itself over a region of parameter space where policy weights sampled according to $P(\theta \mid \mu, \sigma)$ will have high fitness.}
  \label{fig:ES}
  \Description{How evolutionary strategies works - a visualisation }
\end{figure}

However, \citet{Lehman2017} clearly demonstrate that ES is significantly different to finite-difference methods and other point-gradient methods in general as it does not seek to place the mean at the point of highest fitness, but rather centre the entire distribution over a region such that sampling from it leads to solutions with high fitness.  This can lead to unexpected behaviours \cite{Lehman2017}, where the geometry of fitness landscapes can result in the mean parameter returned by ES having significantly lower performance than points nearby it. ES doesn't succumb to the problem of gradient gaps. This is not true for RL.  As ES is required to evaluate the policies that are sampled from the distribution over entire episodes in order to generate gradient updates, a large amount of data is generated.  Other evolutionary approaches such as GAs can be more data efficient in this sense, as it is possible to maintain a smaller population of individuals to evaluate (\textit{pop-size}=10) than ES which can require hundreds of evaluations per gradient update.  While this can make GAs appear more appealing, it has been noted by \citet{Bodnar2020} that GAs can lead to undesirable traits such as catastrophic forgetting of learned behaviours, while \cite{Segal2022} claims GAs fail when a reward signal is either sparse or deceptive as is often the case in control problems.  ES also have the property of directly approximating gradients as opposed to the more heuristic search technique observed in GAs.

\textbf{Evolutionary Reinforcement Learning:}  Evolutionary Reinforcement Learning is a fast emerging area combining RL and EA approaches.  The reason for its appeal is due to the the two approaches having almost directly opposing traits.  RL often follows point gradients which can lead to problems of convergence to local optima, vanishing gradients and gradient gaps.  Conversely, EAs such as GAs are gradient free, while ES do not follow point-gradients resulting in a different behaviour.  Off-policy reinforcement learning algorithms are more data efficient due to their ability to re-use data from the Replay Buffer, however EAs often require the entire population to be evaluated on each generation resulting in much more data.  As such, methods have been proposed which seeks to combine the strengths of the two approaches. 

\citet{Khadka2018} propose a framework in which a GA runs alongside an RL (DDPG) agent.  The GA can feed the RL its excess data by sending its trajectories to the RL agent's replay buffer.  The RL agent can occasionally overwrite the weakest performing member of the GA seeking to improve the overall average fitness of the population.  This leads to a cycle of the RL agent improving the population resulting in better performance during training. These results demonstrate an increase in performance of the compound algorithm over the two halves when they act separately.  

CEM-RL proposed by \citet{Pourchot2018} combines CEM (a method similar to ES, but without directly approximating gradients) with TD3.  CEM maintains a population mean and standard deviation and samples policies for evaluation in a similar way to ES.  However only the top $50\%$ are used in updating both of the population parameters.  In CEM-RL when the sampled policies have been generated, half are evaluated directly while the other half are updated following the TD3 actor update rule for $M$ time steps, before being evaluated.  The algorithm then concludes similar to CEM with the top $50\%$ overall being selected and used to update the distribution parameters.

AES-RL \cite{Lee2020} builds on CEM-RL by introducing a multiple worker asynchronous update framework allowing for the algorithm to benefit from temporal speed-ups and achieving state of the art results in some environments.

CHDRL \cite{zheng2020cooperative} uses a hierarchical architecture to combine a global off-policy RL agent, with a local on-policy RL agent and a local EA (CEM).  This builds on the reasoning that combining EA with RL is good due to their opposing properties, one step further by also trying to make use of the good sample efficiency properties of off-policy RL while also benefiting from the stability properties of on-policy RL.

While ES is not exactly like a finite-difference algorithm as mentioned earlier,  it still closely resembles one.  \citet{shi2019fidi} combine the gradient updates of a RL agent (DDPG) with those of a Finite-Difference algorithm (Augmented Random Search) to update the parameters of one single shared policy network.

ESAC \cite{suri2020maximum} combines an EA with Soft Actor Critic \cite{Haarnoja2018} - a RL algorithm which maximises both the cumulative reward of the agents but also the entropy of the action distributions.  This leads to better exploration of the agent and thus improved results.

The increasing interest in Evolutionary Reinforcement Learning has also lead to the development of platforms to aid researchers and developers implement these algorithms \cite{Bai2022}.

%%%%%%%%%%%%%%%%%%%%%%%%%%%%%%%%%%%%%%%%%%%%%%%%%%%%%%%%%%%%%%%%%%%%%%%%

\section{Methods}
In this section we present our algorithm using a similar framework to ERL\cite{Khadka2018}.  We explain the intuition behind the multibuffer system in this application and provide pseudocode for our proposed algorithm, ES-TD3Buffers.

\subsection{Multi-Buffers}
The framework first presented in ERL allowed for the genetic algorithm to send the trajectories its population of policies generate during their evaluation phase to a single buffer.  The RL agent then appends its own exploratory experience and samples a batch uniformly from the buffer to update its parameters.  

While GAs (where the elite survives from generation to generation) ensures the max score of the population will not decrease over the iterations (in deterministic environments), no such guarantee is observed with ES.  Due to ES following a Monte Carlo gradient approximation, it is easy for it to "fall" from peaks of high fitness if the learning rate is set too high, while being extremely data inefficient if the learning rate is set too small.  Pairing this with its abnormal search strategy which allows for the mean of the Gaussian distribution to centre itself in areas of low fitness provided the surrounding areas have high fitness, runs the risk of the over-production of poor performing trajectories.  Carelessly appending all such trajectories to a single buffer runs the risk of pushing all "good" trajectories out of the buffer and leaving the RL agent with nothing but undesirable behaviours to learn from.  Intuitively, without having any examples of desirable behaviour, the agent runs the risk of learning the best of the undesirable behaviours.

Hence we propose a simple multibuffer approach.  By compartmentalising the replay buffer into "Good","Bad" and "Exploratory" partitions, ES can append all its trajectories without the risk of negatively impacting the RL agent.  In this paper a very simple threshold is used to determine whether a trajectory is considered "Good" or "Bad", namely we track the highest episodic fitness and use $90\%$ of this number as the threshold.  Since the RL agent generates exploratory trajectories on a timestep basis rather than one complete episode at a time, we propose that these data points are stores in the separate compartment as it is unknown if they can be considered "Good" or "Bad" until the episode terminates.  During the learning step the RL agent samples a ratio from each buffer.

\begin{figure}[!b]
    \centering
    \includegraphics[width=\linewidth]{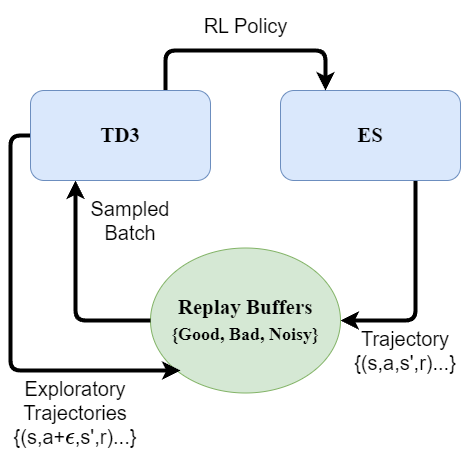}
    \caption{ES-TD3Buffers Framework}{ES begins by generating offspring by sampling from its distribution.  The resulting offspring are evaluated with their trajectories sent to the corresponding compartment of the buffer. 
    TD3 adds data to the noisy buffer when acting in the environment and samples a batch from all three buffers to perform its update.  Periodically ES and TD3 are compared, with the mean of ES being replaced by TD3 if TD3 is outperforming it.}
    \label{fig:ES-TD3Buffers}
    \Description{Visualisation of ES-TD3Buffers,the cycle of communication between evolutionary algorithm and reinforcement learner}
\end{figure}

Buffer augmentation is a technique implemented in previous RL studies \cite{Lv2022}, where a car was trained to safely overtake, a separate buffer was utilised to store any crashes.  As crashing during online training is extremely dangerous and costly, the use of a second buffer minimised the risk of catastrophic forgetting as the policy improved by reminding it periodically of actions that lead to crashes - penalised by heavily negative rewards.  In another recent study, \citet{esmaeeli2022} improved on ERL's results by using an Elite buffer that was generated from the GA's trajectories, but applied data engineering techniques to allow only the most diverse subset of good trajectories be used in the update step. 
%%% should this all be in the background?  but then there would arguably be too much background

\begin{algorithm}[!t]
\caption{ES-TD3Buffers}\label{alg:cap}
\begin{algorithmic}
%\Require $\pi$, $Q$, $\mu$, $\sigma$, $\alpha$, ratio $(a:b:c)$, multibuffer $\beta_G$,$\beta_B$,$\beta_\epsilon$
\State Initialise: TD3 actor $\pi$
\State Initialise: TD3 critics $Q_0$, $Q_1$
\State Initialise: TD3 target critics $\hat{Q_0}$, $\hat{Q_1}$
\State Initialise: ES mean $\mu$ and std $\sigma$
\State Initialise: MultiBuffer: $\beta_G$,$\beta_B$,$\beta_\epsilon$  with sampling ratio $(a:b:c)$
\State Set $Threshold=-\infty$
\For{$\infty$}
\For{M frames} \Comment{perform TD3 iterations}
\State Reset env $s=s_0$
\State $ep\_reward=0$
\While{Episode not terminated}
\State $a \gets \pi(s) + \epsilon$  \Comment{Generate exploratory action}
\State take action $a$ in env, add $(s,a,s',r)$ to $\beta_\epsilon$
\State $ep\_reward$ +=$r$
\If{All buffers contain K datapoints}
    \State Sample batch from $\beta_G$,$\beta_B$,$\beta_\epsilon$, under ratio$(a:b:c)$ 
    \State Update $\pi$ and $Q_0$, $Q_1$ using TD3 update rule
    \State Periodically soft-update $\hat{Q_0}$, $\hat{Q_1}$
\EndIf
\EndWhile
    \If{$ep\_reward > Threshold$}
        \State $Threshold \gets ep\_reward$
    \EndIf
\EndFor  %the end for the td3 part
\State $TD3 \gets Evaluate(\pi)[0]$
\For{g Generations}  \Comment{perform ES iterations}
\For{$i$ in n} \Comment{For each ES offspring}
\State Sample noise $N_i \sim \mathcal{N}(0,\,\sigma^{2})$
\State Generate offspring $X_i=\mu+N_i$
\State $F_i,trajectory=Evaluate(X_i)$
\If{$F_i > 0.9*Threshold$}
    \State Send $trajectory$ to $\beta_G$
    \If{$F_i > Threshold$}
        \State $Threshold \gets F_i$
    \EndIf
\Else{}
\State Send trajectory to $\beta_B$
\EndIf
\EndFor
\State $\mu \gets \frac{1}{n}\sum_{i=1}^{n} F_i N_i $
\State $ES \gets Evaluate(\mu)[0]$
\EndFor %the ES part
\If{$TD3>ES$}
    \State $\mu \gets \pi$
\EndIf
\EndFor %the iterations
\State \textbf{Return:} $\mu$

\end{algorithmic}
\end{algorithm}

\subsection{Overwrite Rule}
In ERL, the use of a GA allowed for the RL policy to be periodically substituted for the lowest performing individual in the population.  Since ES can be understood as maintaining a population of size one, namely the policy described by $\mu$, any attempt to insert the RL policy will result in a greater loss of information than the GA setting.  In this paper we utilise a simple overwrite rule such that if the RL policy is performing better than the ES mean, then the ES mean should be replaced with the RL policy parameters.  In practice this was done by averaging the episodic scores of RL and ES over a number of iterations as these algorithms can occasionally significantly decrease in performance between updates. Utilising averages will mitigate such effects in the comparison.

After implementation we found that ES often settled on local optima,  the use of the overwrite rule can be interpreted as providing a means for ES's distribution to be re-centred over the search space of parameters if the RL agent finds an area with higher performance.  This then will allow ES to generate offspring in this "fitter" region of space and hence avoiding the replay buffers being filled with the same low fitness trajectories for the remaining run time.

\begin{algorithm}[t]
\caption{Evaluate}\label{alg:cap2}
\begin{algorithmic}
\Require Env, policy $\pi$
\State Reset Env $s \gets s_0$
\State $Fitness \gets 0$
\State $Trajectory = []$
\While{Env not terminated}
\State $a \gets \pi(s)$
\State Take action $a$ in env, and observe reward $r$ and next state $s'$
\State Append $(s,a,s',r)$ to $Trajectory$
\State $Fitness \gets Fitness+r$
\State $s \gets s'$
\EndWhile
\State \textbf{Return:}$Fitness, Trajectory$
\end{algorithmic}
\end{algorithm}

\subsection{ES-TD3Buffers}
To implement our algorithm, a TD3 actor $\pi$, dual critics $Q_0$, $Q_1$ and corresponding target critics are initialised.  Here we use Neural Networks as function approximators for each.  Additionally an isotropic multivariate Gaussian distribution with mean $\mu$ and standard deviation $\sigma$ is centred over the origin of the space of all parameters of the actor network.  Furthermore, a "MultiBuffer" consisting of 3 empty buffers is initialised.  

Periods of TD3 updates are applied to $\pi$, $Q_0$ and $Q_1$ (with soft updates applied to the target critics). This is followed by a fixed number of generations of sequential updates to the search distribution parameter $\mu$, as described in \citet{Salimans2017}.  The two aforementioned processes are then iterated until a stopping criteria has been met (here fixed number of iterations).
During each generation of ES, a population of actors are evaluated, which generates trajectories of data.  These data points are sorted into the correct replay buffer ("Good" or "Bad") by comparing the reward gathered over the entire episode to the current "threshold".  The simple "threshold" used in this paper is $90\%$ of the highest recorded ES episode to date.
TD3 on the other hand, generates "noisy" data when selecting actions during training by adding Gaussian noise to the TD3 policy.  As such we place this data in the third replay buffer.
After TD3 performs a "noisy" action in the environment, a batch of data is drawn from the three buffers following a predefined ratio (see table \ref{tab:hyperparameters}).
Gradient ascent is performed on the parameters of TD3s actor and critics as first described in \citet{Fujimoto2018}.

 After every iteration a comparison is made between the TD3 actor and the policy defined by $\mu$.  If the TD3 actor $\pi$ is outperforming $\mu$, we re-centre the ES search distribution over $\pi$.  Pseudocode for this can be found in Algorithm \ref{alg:cap}.

\section{Experimental Study}
In this section we evaluate the performance of the ES-TD3Buffers algorithm.  In particular we aim to answer the following questions:
\begin{itemize}
    \item How does ES-TD3Buffers perform in control tasks compared to pre-existing Evolutionary Reinforcement Learning algorithms?
    %\item What empirical advantage does the multi replay buffer offer over a single buffer approach?
    \item What limitations does the algorithm have?
\end{itemize}

\subsection{Environments}
We used the well known OpenAI gym package which offers a variety of continuous state and action space environments through the MuJoCo physics simulator.  

The majority of environments consist of controlling a multi-jointed robot with the goal of performing a task such as running as fast as possible or staying upright for as long as possible.  The action space consists of the set of all force vectors that can be applied to the joints.  

\begin{figure*}[h]
    \centering
    \includegraphics[width=\linewidth]{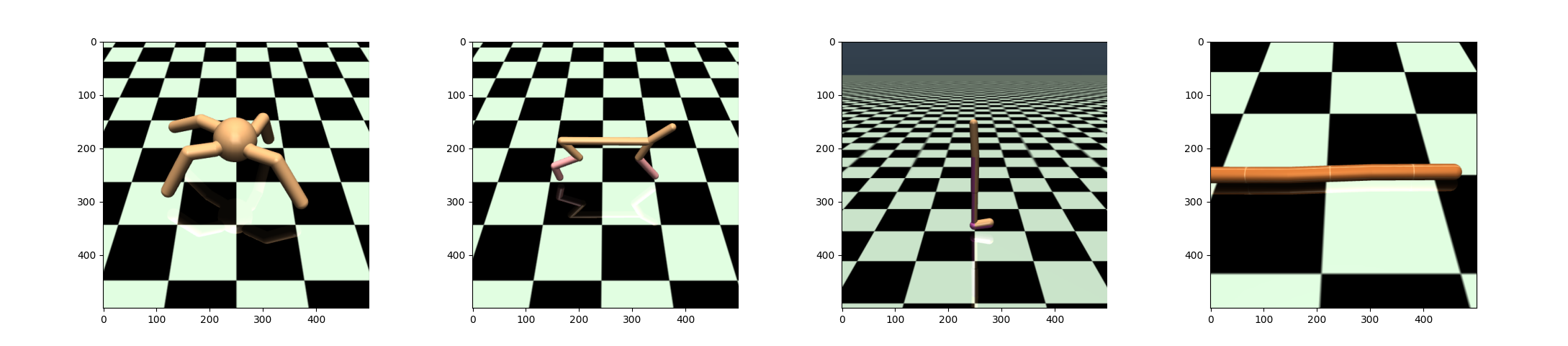}
    \caption{MuJoCo Environments}{From left to right: Ant,  HalfCheetah, Walker and Swimmer.  The goal in all environments is for the creature to learn to move at a high velocity.}
    \label{fig:mujoco}
    \Description{MuJoCo environments used in analysis}
\end{figure*}

In our experiments we used "HalfCheetah", "Swimmer", "Ant" and "Walker2d" as they are commonly analysed environments in this area. Figure \ref{fig:mujoco} illustrates these problem domains.

\subsection{Architecture}
In our implementation we used feed-forward Neural Networks to represent all actors and critics.  The actors consisted of 2 hidden layers and 256 neurons per hidden layer.  The number of output nodes was equal to the dimension of the action space for that particular environment, similarly the number of input neurons was equal to the dimension of the state space.  The networks used hyperbolic tangent activation functions between all layers.  

The critic consisted of 2x256 hidden layers, while the input layer's number of neurons was set to the sum of the state space dimensionality plus the action space dimensionality. The output layer consisted of a single neuron.  The learning rate for the TD3 actor and critic was set to 0.0003, while the ES used a learning rate of 0.001.  For more details see Table \ref{tab:hyperparameters}.

\subsection{Results}
We compare our algorithm to several well known and commonly used RL and EC control algorithms in Table \ref{tab:baselines} .  

TD3 is reported as it is used as a building block within our ES-TD3Buffers algorithm.  CEM is a common alternative to ES which follows a very similar update rule where we only use the fittest K offspring to update our search distribution as opposed to using all offspring as in ES. Finally ERL and CEM-RL are reported as two of the most influential examples of Evolutionary guided reinforcement learning.

The ES-TD3Buffers is ran for a total of 20 iterations, where each iteration consists of 50 updates to the ES distribution mean,$\mu$, 100'000 TD3 timesteps and a chance for $\mu$ to be overwritten with $\pi$.

All baselines shown are those reported in \citet{Lee2020}.

\begin{table*}[!h]
    \large
    \caption{Scores achieved on MuJoCo Environments. Baselines As Reported in \citet{Lee2020}}
    \label{tab:baselines}
    \centering
    \begin{tabular}{cc|c|c|c|c|c}\toprule
    \textit{Environment} & \textit{Statistics} & \textit{TD3} & \textit{CEM} &\textit{ERL} &\textit{CEM-RL} &\textit{ES-TD3Buffers}\\ \midrule
     & Mean & 9630 & 2940 & 8684 & 10725 &\textbf{10793}\\
     HalfCheetah & Std. & 202 &353&130&397& 778\\
     & Median & 9606 &3045&8675 &11539& 10862\\ \bottomrule
    & Mean & 4027 & 487&3716 & 4251& \textbf{4532}\\
    Ant & Std. & 403 & 33 &673& 251& 999\\
    & Median & 4587  &506 &4240 &4310& 4367\\ \bottomrule 
    &Mean & 63 &\textbf{351}&350 & 75& 213\\
    Swimmer & Std. & 9 &9&8 &11& 119\\
    & Median & 47 &361&360 &62& 174\\ \bottomrule
    & Mean & 3808 & 928  & 2188  & \textbf{4711}   & 2217 \\
    Walker & Std. & 339 &50 & 240&155  & 1454\\
    & Median & 3882 & 934 &2267 &4637 & 1764\\ \bottomrule \bottomrule
  \end{tabular}
    \label{tab:baselines}
\end{table*}

In the interest of fairness and clarity, we note that ES-TD3Buffers uses significantly more data than other reported algorithms.  This is primarily due to the fact that ES requires significantly more data than some other EAs including GAs and CEM as to approximate the gradients in high dimensional spaces we often require the number of offspring used to be in the range of hundreds to thousands.

We claim the increase in samples should not falsify ES-TD3Buffers claim as a competitive algorithm under the reasoning provided by Salimans in his original paper \cite{Salimans2017}.  Namely since ES does not require backpropagation to update the parameters of the distribution, it can run in competitive time to other more data efficient gradient-based algorithms.  In the future works section, we discuss how parallel computing could be leveraged to speed this up further.

For comparative purposes, we ran our algorithm such that the TD3 half generates 2 million frames.

The results show ES-TD3Buffers performs comparably with the state of the art - CEM-RL, across a variety of control tasks.  Notably ES-TD3Buffers is able to outperform CEM-RL on both HalfCheetah, Ant and Swimmer, while performing comparatively to ERL on Walker.

We note that while the average performance of our algorithm is high, the standard deviation is also.  We expect that this is caused in part to the overwrite rule, which can take the small variance in TD3 runs -especially at the early stages - and choose to make vastly different overwrites to the ES actor, which will ultimately affect the quality of data TD3 is trained on over the subsequent iterations.  A possible solution to this which we will investigate is to implement a softer-update rule.

Furthermore, we note that we were unable to achieve the same results with our version of TD3 on the Walker environment as those reported in \citet{Lee2020} and shown in Table \ref{tab:baselines}.  The Github containing the TD3 source code, which we used, does state hyperparameters have been changed from the original implementation and we expect that this could be the reasoning for why our ES-TD3Buffers does not perform at least as well as the reported TD3 results on the Walker environment.

%%add a note saying td3 didnt perform as well as 

\subsection{ES vs TD3 Learning Dynamics}
\begin{figure}[h]
    \centering
    \includegraphics[width=\linewidth]{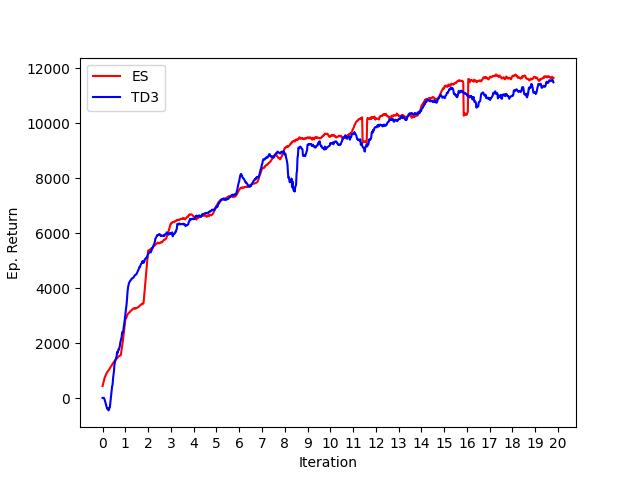}
    \caption{ES and TD3 HalfCheetah Learning Curves}{ES and TD3 both run for 20 iterations of their own update rule, and on completion of each iteration, a performance check is carried out which can lead to $\mu$ being overwritten with $\pi$.  The curves shown are smoothed with a moving window of size 10 to provide a clearer visual aid}
    \label{fig:learning Curves}
    \Description{Showcasing how the two halves of the algorithm learn separately and where they are combining}
\end{figure}

In this section we examine how TD3 and ES work together in ES-TD3Buffers to improve on the results they achieve individually across the majority of MuJoCo environments.
Figure \ref{fig:learning Curves} shows the individual learning curves for both the ES and TD3 halves of our algorithm, across the full 20 iterations on the HalfCheetah environment.  An iteration consists of 50 generations of ES and 100'000 TD3 timesteps.  
The figure showcases the typical behaviour witnessed across the other environments - Walker and Ant, however not Swimmer for reasons we discuss later.  
Most notably Figure \ref{fig:learning Curves} allows us to see how ES quickly converges to a local optimum by the end of the 1st generation.  In our experiments with ES we found it seldom escaped these local optima for a wide choice of hyperparameters.  This is evident by the results of the closely related CEM in Table \ref{tab:baselines}, which performs poorly across HalfCheetah, Ant and Walker when compared to the other algorithms.
By allowing TD3 to overwrite the mean $\mu$ of the Evolutionary Strategies search distribution with its own actor $\pi$, we see how ES gets lifted out of the local optima of the early generations.
Most interestingly is that while ES struggles at the beginning of training on these environments, it usually can be seen to exceed the converged TD3 performance over the last few iterations (iterations 14 to 20 in Figure \ref{fig:learning Curves}).
In the case of Swimmer,  ES is capable of achieving much better results than TD3 when ran individually, as seen again by the closely related CEM achieving the best results on this environment in Table \ref{tab:baselines}.  Thus there are no early overwrites performed by TD3 in that environment.

\begin{table*}[!t]
    \large
    \caption{Hyperparameters}
    \label{tab:hyperparameters}
    \centering
    \begin{tabular}{c|c|c}\toprule
    \textit{Hyperparameter} & \textit{Description} & \textit{Value} \\ \midrule
     $\sigma$ & ES Standard Deviation & 0.005\\
     $\mu$  & ES Mean Vector & Randomly initialised near 0\\
     $\alpha_{ES}$& ES Learning Rate & 0.001 \\ 
     $n$ & Number of ES Offspring & 60 \\ \bottomrule
     $\alpha_{TD3}$ & TD3 Learning Rate & 0.0003 \\
     $\epsilon$ & Exploratory Noise Added To TD3 & Sampled from $\mathcal{N}(0\, 0.1)$\\
     $\tau$ & TD3 soft update hyperparameter & 0.005 \\
     K & Min Number of TD3 Exploratory Timesteps Before Learning Starts & 25000 \\ \bottomrule 
    (a,b,c)& Sampling Ratio of Good, Bad and Noisy DataPoints & (0.5,0.2,0.3) \\
    $Threshold$ & Threshold for trajectory being appended to "good" buffer & 0.9x(highest recorded fitness) \\
    M & Number of TD3 Frames Between ES Iterations & 100000\\
    g & Number of ES Generations Between TD3 Iterations & 50\\
     \bottomrule\bottomrule
  \end{tabular}
    %\label{tab:baselines}
\end{table*}

\section{Discussion}
We have demonstrated that Evolutionary Strategies is a highly effective evolutionary method for ERL algorithms. The results presented in this paper contribute to the existing literature in multiple respects. \\ 
\balance
Firstly, as reported in the Results section,  ES is significantly more data inefficient than GAs.  This is due to the fact that GAs can be run effectively with population sizes of $\le10$.  On the contrary, using Neural Networks as function approximators in ES often causes the dimensionality of the search space to be very large.  As ES is seeking to Monte Carlo approximate the gradient in a method similar to finite-differencing, the number of samples required in high dimensional spaces should also be large.  

In practice we found ES required the creation of 60 offspring per generation to achieve good performance.  Since antithetic sampling is implemented, this is equivalent to 120 offspring in the population.  This already results in $10x$ more data being used than in the GA case.  Methods such as E-ERL \cite{Wu2022} have been proposed which reduces the data usage of ERL by only allowing the GA component of the algorithm to run when the RL component has converged to a local optimum.  Such a technique could also reduce the data usage when ES is used in place of a GA as per our algorithm.

Secondly, the overwrite rule in this paper can result in a huge loss of information in the ES distribution.  In the case of a genetic algorithm, only a single member of the population is overwritten, but due to the population being represented by the mean parameter in ES the population can be represented as having a cardinality of 1.  Alternative overwrite rules have been explored such as that of \citet{Jung2020}, which avoids convergence to the same areas of the search space by implementing a "soft update".  A soft update in ES-TD3Buffers could push the ES actor towards the RL in parameter space, without setting it directly equal to the RL actor.

Thirdly, the original ES paper \cite{Salimans2017} demonstrated experimentally how efficiently ES can run across parallel workers.  The parallel version of ES reduced training times of some MuJoco problems from 18 hours down to 10 minutes.  As such, the excess of required data could be justified.  Due to the way in which our algorithm requires trajectories to be shared between the ES offspring and the RL agent, our algorithm cannot be parallelised in the exact way in which \citet{Salimans2017} did.

Hogwild! \cite{hogwild2011} is a platform which gained popularity in RL \cite{Mnih2015} by allowing a central reinforcement learner to share its parameters amongst several parallel workers, and in return asynchronously receive gradient updates it should perform on the central parameters from each worker.  A platform similar to this could be used in the case of ES-TD3Buffers to store the RL agent and ES distributional parameters on the central worker while allowing ES to evaluate its offspring in parallel, communicating back the trajectories with the gradients.

Lastly, NSRA-ES \cite{Conti2018} changed the definition of fitness in ES from purely episodic score to a linear combination of episodic score and novelty with respect to some archive of previously encountered behaviours.  The coefficients of this linear combination could adaptively change weight to further emphasise optimising episodic score when improvements were being made, while increasing the emphasis on finding new behaviours when the episodic scores stopped improving.  NSRA-ES would be a perfect substitute for ES in ES-TD3Buffers and further encourage the discovery of more diverse behaviours being added to the replay buffer.

\section{Conclusion}
In this paper we have presented a new algorithm called ES-TD3Buffers.  The ES-TD3Buffers algorithm demonstrates that Evolutionary Strategies is highly effective when combined with RL algorithms for Evolutionary Reinforcement Learning. A portion of the success of ES-TD3Buffers is due to its unique multi-buffer architecture. Our algorithm performs comparably to the current state-of-the-art Evolutionary Reinforcement Learning algorithm (CEM-RL) when tested on MuJoCo enviornments. ES-TD3Buffers provides a improvement of $184\%$ on Swimmer, a $6.6\%$ improvement on Ant and provides competitive performance on HalfCheetah when compared directly to CEM-RL.
Our algorithm also shows that ES with TD3 works better in this compounded framework than they do separately, as our algorithm outperforms TD3 on 3 of the 4 environments tested, with reasoning given in section 4.3 for the failure to improve on Walker. This is in thanks due to the RL agent helping ES escape poor local optima early in the training, which ultimately leads to better training data for the RL agent.

\begin{acks}
This work was conducted with the financial support of the Science Foundation Ireland Centre for Research Training in Artificial Intelligence under Grant No. 18/CRT/6223
\end{acks}

%%%%%%%%%%%%%%%%%%%%%%%%%%%%%%%%%%%%%%%%%%%%%%%%%%%%%%%%%%%%%%%%%%%%%%%%

%%% The acknowledgments section is defined using the "acks" environment
%%% (rather than an unnumbered section). The use of this environment 
%%% ensures the proper identification of the section in the article 
%%% metadata as well as the consistent spelling of the heading.

%%%%%%%%%%%%%%%%%%%%%%%%%%%%%%%%%%%%%%%%%%%%%%%%%%%%%%%%%%%%%%%%%%%%%%%%

%%% The next two lines define, first, the 
%bibliography style to be 
%%% applied, and, second, the bibliography file to be used.

\bibliographystyle{ACM-Reference-Format} 
\bibliography{bib}

%%%%%%%%%%%%%%%%%%%%%%%%%%%%%%%%%%%%%%%%%%%%%%%%%%%%%%%%%%%%%%%%%%%%%%%%

\end{document}